\documentclass{article}
\usepackage[english]{babel}
\usepackage[a4paper,top=2cm,bottom=2cm,left=3cm,right=3cm,marginparwidth=1.75cm]{geometry}
\usepackage{amsmath}
\usepackage{graphicx}
\usepackage[hidelinks]{hyperref}
  \usepackage[
    backend=biber,
    style=numeric,
  ]{biblatex}

\title{I Was Blind but Now I See: Implementing Vision-Enabled Dialogue in Social Robots}
\author{
    Giulio Antonio Abbo and Tony Belpaeme\\
    \textit{IDLab-AIRO -- Ghent University -- imec, Belgium}\\
    \small\texttt{giulioantonio.abbo@ugent.be}
}
\date{\scriptsize arXiv preprint 15 Nov 2023}

\begin{document}
\maketitle

\begin{abstract}
    In the rapidly evolving landscape of human-computer interaction, the integration of vision capabilities into conversational agents stands as a crucial advancement. This paper presents an initial implementation of a dialogue manager that leverages the latest progress in Large Language Models (e.g., GPT-4, IDEFICS) to enhance the traditional text-based prompts with real-time visual input. LLMs are used to interpret both textual prompts and visual stimuli, creating a more contextually aware conversational agent. The system's prompt engineering, incorporating dialogue with summarisation of the images, ensures a balance between context preservation and computational efficiency. Six interactions with a Furhat robot powered by this system are reported, illustrating and discussing the results obtained. By implementing this vision-enabled dialogue system, the paper envisions a future where conversational agents seamlessly blend textual and visual modalities, enabling richer, more context-aware dialogues.
\end{abstract}

\section{Introduction}

In the ever-evolving landscape of human-computer interaction, the quest for more intuitive and immersive experiences has fueled advancements in natural language processing and artificial intelligence.
Conversational agents, ranging from chatbots to robots, play a pivotal role in this evolution.
As we witness the increasing integration of these agents into our daily lives -- from home assistants to help desks, and from elderly care to teaching -- the demand for richer, context-aware conversations becomes more pronounced.

While Large Language Models (LLMs) have demonstrated remarkable abilities in generating human-like text, their traditional reliance on purely textual inputs leaves a gap in achieving a holistic understanding of the user's context.
As conversations unfold, individuals naturally incorporate visual cues, expressions, and environmental context to enhance communication.
Consider a scenario where words alone cannot capture the nuances of a conversation; a visual element could provide the missing link.
With the advent of image input for LLMs, the challenge lies in orchestrating a seamless integration of textual and visual information.

This paper addresses precisely this gap by leveraging the new vision capabilities of LLMs, allowing conversational agents not only to decipher textual inputs but also assimilate and respond to visual stimuli in real-time.
The integration of visual elements, captured through frames from a live video feed, enhances the agent's contextual awareness, fostering a more natural and immersive conversational experience.
The frame summarisation process presented allows balancing the trade-off between the amount of visual information and the computational cost of processing it.
We present an initial implementation\footnote{\url{https://github.com/giubots/vision-enabled-dialogue} -- A release of the code at the time of writing is published at \url{https://zenodo.org/doi/10.5281/zenodo.10127384}.} of this vision-based conversational agent, which can be used both as a standalone application and to power a Furhat robot~\cite{al_moubayed_furhat_2012}.

\section{Background}

\subsection{Large Language Models for Dialogue Generation}

Large Language Models (LLMs) are advanced natural language processing models that leverage deep learning techniques, particularly neural networks, to understand and generate human-like text.
These models, like GPT-4~\cite{openai_gpt-4_2023} or LLaMA-2~\cite{touvron_llama_2023}, are trained on massive amounts of diverse textual data, allowing them to learn intricate patterns, syntactic structures, and semantic relationships present in language.
LLMs operate by transforming words into high-dimensional vectors, enabling them to capture and generate contextually relevant and coherent text based on input prompts.

In the context of chatbots and dialogue systems, LLMs play a crucial role in enhancing conversational experiences.
These models are capable of generating text-based responses that not only capture the intricacies of human language but also demonstrate a nuanced understanding of the context.
As a result, the model has common ground with users, both simulating coherent and contextually relevant conversations and exhibiting traits such as politeness, empathy, and genuine interest in the interaction.
This versatility makes LLMs invaluable for applications like conversational agents, facilitating the creation of more engaging and natural dialogues.

The output of an LLM depends mainly on the input instructions, that is, the prompt used.
Prompt engineering is the process of finding the best prompt for the desired output, and there are many techniques.
Of these, one relevant to our work is using an additional LLM to summarise parts of the dialogue history.
Indeed, a prompt that implements a conversational agent contains -- after some initial instructions to guide the quality of the responses -- all the lines of the conversation, so that the LLM has a memory of what has been said.
Parts of this dialogue can be summarised, reducing the length of the prompt while maintaining the context.
This and other approaches have contributed greatly to the quality of the interactions with the conversational agents.

Until recently, the context awareness of the responses relied on a textual description of the contextual information~\cite{zeng_socratic_2022, wang_egocentric_2023}.
For example, Janssens et al.~\cite{janssens_cool_2022} propose a system where a captioning model describes an image and another model is used to generate appropriate text based on that caption.
A significant advancement has been the incorporation of non-textual information, such as images, into the input of these models (e.g., GPT-4, IDEFICS~\cite{laurencon_obelics_2023}).
Thanks to this development, LLMs can now analyse and generate text based on visual cues, enabling a more comprehensive understanding of the world.
This evolution opens up new possibilities for context-aware applications, in which a combination of textual and visual information is essential.

Visual language models face some real challenges.
First, they tend to give too much importance to the pictures in the prompt, thus the model starts describing the contents in detail, instead of using the picture for context.
In addition, when mixing dialogue and images in the input, the models tend to get sidetracked by the pictures and lose the thread of the conversation.
Finally, the increased amount of data to process leads to increased computing time, with a sometimes considerable delay in the generation of the answer.
Tackling these issues is crucial to making visual elements work seamlessly in chat and conversation systems.

\subsection{Measures for Conversational Interactions}

In the field of human-computer interaction, researchers have mostly been interested in the pragmatic use of language: as an interface to tools to achieve a goal.
Other uses, like information, persuasion, entertainment and social bonding, lagged behind as they were difficult to achieve.
Indeed, programs struggled to pass the Turing test and their users were always aware that their interlocutor was a just cold machine.
Today, increasingly sophisticated algorithms and the advent of LLMs bring us close to an actual suspension of disbelief as we take part in conversations with agents and robots. 
To continuously improve the quality of the conversation it is useful to understand what makes a \emph{good} conversation.

A focus group on conversational agents~\cite{clark_what_2019} revealed that \emph{people fundamentally questioned the need for bond and common ground in agent communication, shifting to more utilitarian definitions of conversational qualities.}
Considering human-human and human-machine interactions the study identifies the following requirements:
\emph{mutual understanding} and \emph{common ground} which, when it comes to machines, are translated to remembering preferences and facts about the user;
\emph{trustworthiness}, which becomes \emph{privacy} in the machine case, a concern that emerged also in other works~\cite{abbo_users_2023};
\emph{active listening}, and \emph{accurate listening} with digital assistants, i.e. reducing the need to repeat oneself;
\emph{humour}, which is a welcome novelty in human-machine interaction but must have substance and relevance to the conversation.

The TRINDI checklist~\cite{bohiln_survey_1999} offers a way of evaluating the capabilities of a task-oriented conversational interface.
Among the other features, it considers whether the system can adapt to the information provided by the user, for example in handling missing or inconsistent information, barge-in input and no answers (turn-skipping).
In addition, the list also considers whether the system is context-aware in its answer's interpretation and follow-up questions.
However, when it comes to conversational chatbots, users' acceptance is influenced by other qualities: namely, Politeness, Entertainment, Attentive Curiosity, and Empathy~\cite{svikhnushina_key_2021}.

When considering embodied conversational agents many features can influence the relationship quality~\cite{loveys_effect_2020}.
Of these, we mention again \emph{humour}~\cite{kulms_lets_2014}, \emph{social reasoning language}~\cite{romero_cognitive-inspired_2017} -- such as acknowledgement, praise, and questions to elicit self-disclosure -- and context awareness~\cite{bickmore_context_2009}.

The advent of LLMs allows building agents that satisfy some of these requisites, for example, common ground, showing respect, empathy, and accurate listening, while others remain an open problem.
We choose to consider the following aspects as the currently most relevant and still problematic:
(a) Personalisation, adapting to the history and previous conversations with the users, while respecting their privacy.
(b) Active listening, especially in showing receptivity and understanding.
(c) Adaptability, both to unexpected contents (unexpected, missing, inconsistent information) and to unexpected flow (barge-in, turn-skipping).
(d) Relevance of the dialogue, in terms of context-awareness and salience to the conversation.
(e) Entertainment and humour in the answers.

\section{Implementing Vision-Enabled Dialogue}

The system proposed empowers a conversational agent with vision capabilities.
When a user interacts with the system, the responses will be grounded in reality and aware of the context, thanks to additional visual input.
This visual input consists of frames from a video captured as the conversation takes place, which are weaved into the conversation.
The LLM that produces the output is instructed to interpret these images as its own sight sense.
For this implementation, we chose GPT-4 as the underlying LLM, as it offers a good balance between costs and performance.
While the system can work on its own, using a webcam and the terminal's text interface, to make the demonstration more realistic we have chosen to use a Furhat robot.

\subsection{Components and Implementation}

\begin{figure}
\centering
\includegraphics[width=\linewidth]{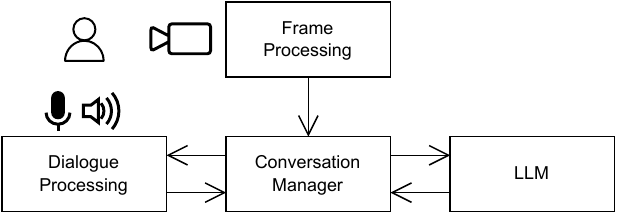}
\caption{\label{fig:components}Overview of the components of the system: the dialogue manager receives inputs from the frame and dialogue processing components, and uses a LLM to produce the outputs.}
\end{figure}

The system implemented is composed of four components as shown in Figure~\ref{fig:components}: the frame and dialogue processing components, the conversation manager and an external LLM.

The \emph{frame processing} component is in charge of retrieving the frames from a video feed and sending them to the conversation manager.
The frequency with which the frames are sent can be configured.
Currently, due to technical limitations, the frame rate must be very low.
We have found that a good compromise between speed and conversation quality is one frame every five seconds.
The component is implemented in three variants: using the video feed from the built-in camera of the Furhat robot, using a webcam, or using a video file.

The \emph{dialogue processing} component provides the user input to the conversation manager and shows the output back to the user.
This component runs in parallel with the previous one, meaning that the conversation manager can receive dialogue and frame inputs in any order.
In our implementation, the dialogue is user-initiated.
There are three implementations available for this component: a text-based input using the terminal, a file-based input for testing purposes, and the Furhat implementation.
The Furhat implementation instructs the robot to look at the user and uses the built-in speech-to-text and end-of-speech detection capabilities of the robot to obtain input from the user.
While the input is being elaborated, the robot looks away, to signal that the robot is not listening anymore.
When the result is ready, the robot looks again at the user and tells the answer leveraging the text-to-speech module, which also controls the mouth movements.

The \emph{conversation manager} is the most important component, as it is in charge of managing the prompt that generates the responses of the system.
When a frame or a message from the user is received, they are added to the prompt, and in the case of messages, a response is generated using the LLM.
For this implementation, the LLM used is GPT-4.
In addition to this, the conversation manager summarises the frames to reduce the prompt length.

\subsection{The Prompt}

The prompt initially consists of a list of frames and dialogue lines, preceded by the following instructions.
\emph{You are impersonating a friendly kid. 
In this conversation, what you see is represented by the images. 
For example, the images will show you the environment you are in and possibly the person you are talking to. 
Try to start the conversation by saying something about the person you are talking to if there is one, based on accessories, clothes, etc. 
If there is no person, try to say something about the environment, but do not describe the environment! 
Have a nice conversation and try to be curious! 
It is important that you keep your answers short and to the point. 
DO NOT INCLUDE EMOTICONS OR SMILEYS IN YOUR ANSWERS.}

Impersonating a kid has been found to improve the quality of the answers, reducing unwanted messages about the capabilities of the model and other disclaimers.
Then, the prompt tells the model how to interpret the images in the prompt, we found this wording to be the most effective so far, compared to more technical explanations.
The rest of the sentences are necessary to reduce the loquacity of the system, and to keep the output relevant and salient.

\subsection{Frames Summarisation}

Continuously adding frames to the prompt leads to an undesirable increase in its size, with longer computation times and costs.
To shorten the prompt we propose to summarise the frames, in a similar fashion to what several implementations do with the conversation.

A naive solution would be to send the first part of the dialogue and frames to a LLM and ask for a summary.
However, this would impact negatively on the conversation quality for three main reasons.
First, as the saying goes, ``an image is worth a thousand words'' and thus the majority of the summary would be devoted to a description of the frames, leaving less room to a summary of the conversation.
Second, this problem would be made even worse with high frame rates, as there would be considerably more pictures than dialogue lines in the prompt.
Third, since the focus of the interaction is the dialogue, it does not make sense to summarise this with the frames, which serve only as context.

\begin{figure}
\centering
\includegraphics[width=\linewidth]{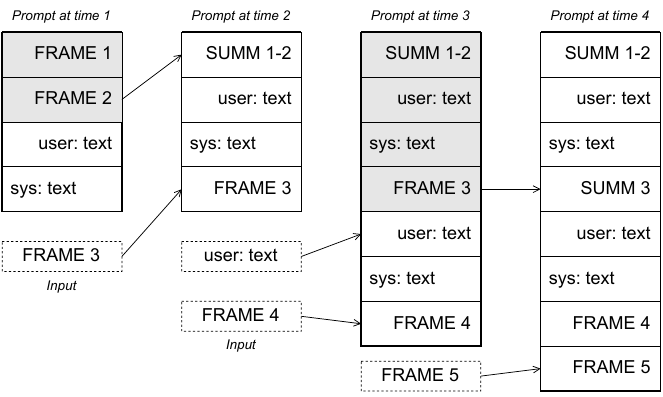}
\caption{\label{fig:summarisation}Example of the summarisation process. Considering $n=3$ and $m=2$. In the first step, a frame is added. The number of frames is now $n$, so the algorithm summarises the first two. Then a dialogue line is added, with the system response, and then another frame is added. In the final step, \texttt{FRAME 5} triggers another summarisation. This time, only \texttt{FRAME 3} is summarised, as including the following frame would disrupt the ordering of the elements. In grey the elements used to obtain the summary at each step: notice that the previous part of the conversation is included.}
\end{figure}

Our solution summarises the frames separately and keeps the ordering of the frames and dialogue lines.
To achieve this, when a frame is received, the conversation manager checks how many frames are in the prompt, if a configurable limit $n$ is reached it performs a summarisation routine.
This routine will scan the prompt, and summarise the first $m$ consecutive frames, as shown in Figure~\ref{fig:summarisation}.
Setting $m<n$ ensures that at least one frame remains in the prompt, to maintain context-awareness.
We have found that keeping at most $n=4$ frames in the prompt and summarising in chunks of $m=3$ frames yields satisfying results.

To obtain the summary, an LLM is prompted with the full conversation and previous summaries up to the frames to summarise and is asked to provide a brief description.
The frames are then removed from the conversation and substituted by their summary.

\section{Interactions and Discussion}

\begin{figure}
\centering
\includegraphics[width=0.45\linewidth]{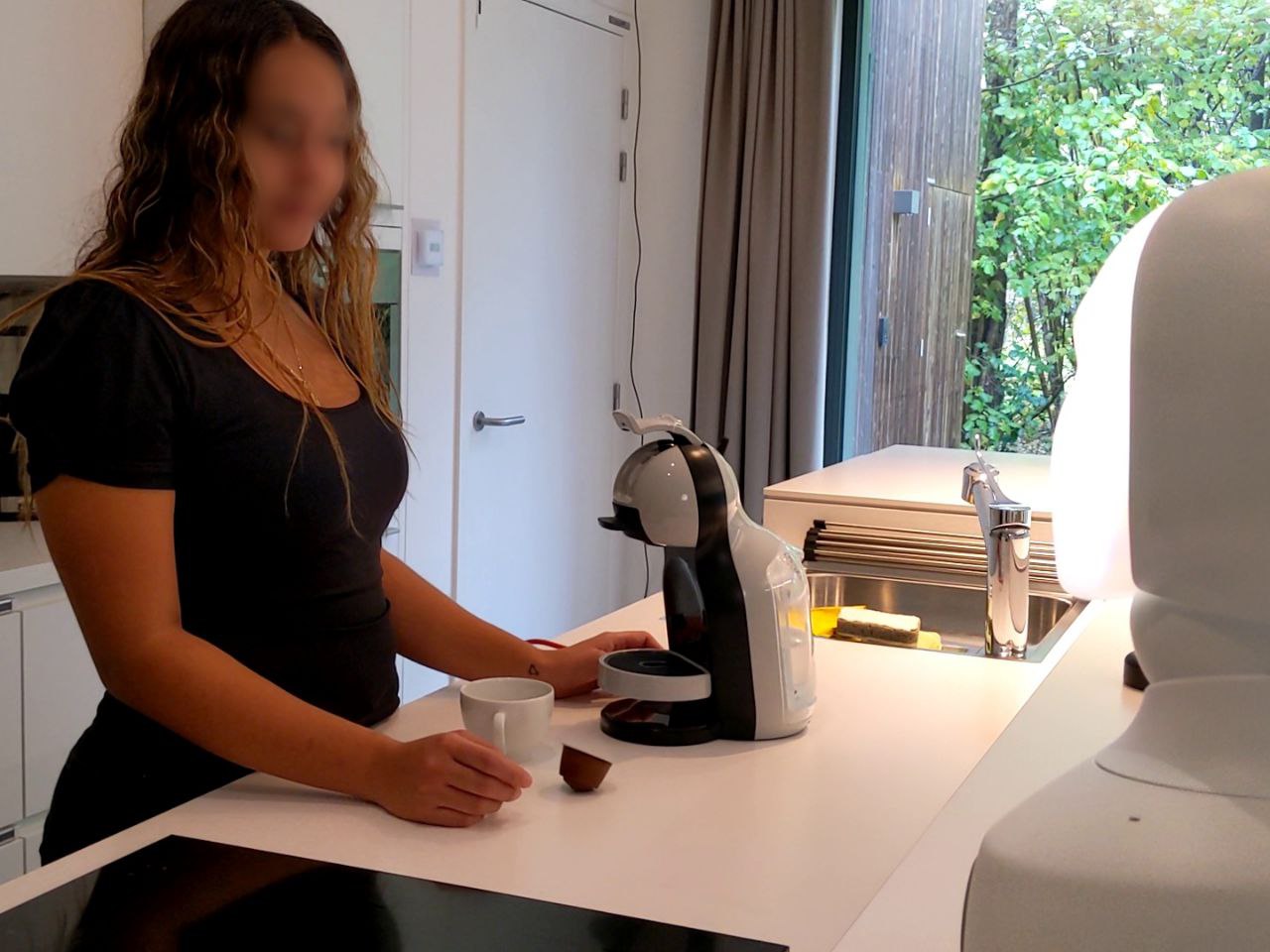}
\includegraphics[width=0.45\linewidth]{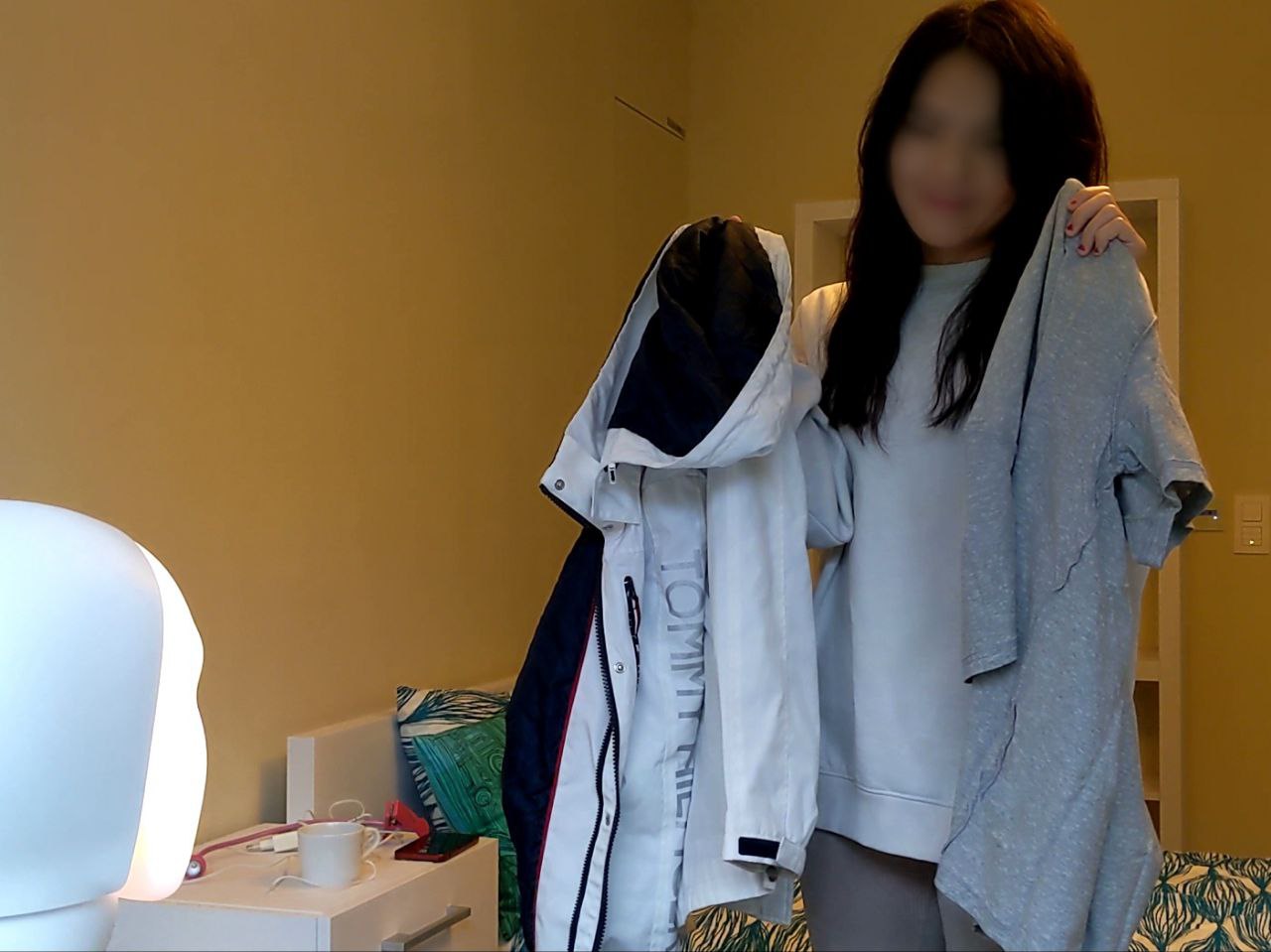}\\
\includegraphics[width=0.45\linewidth]{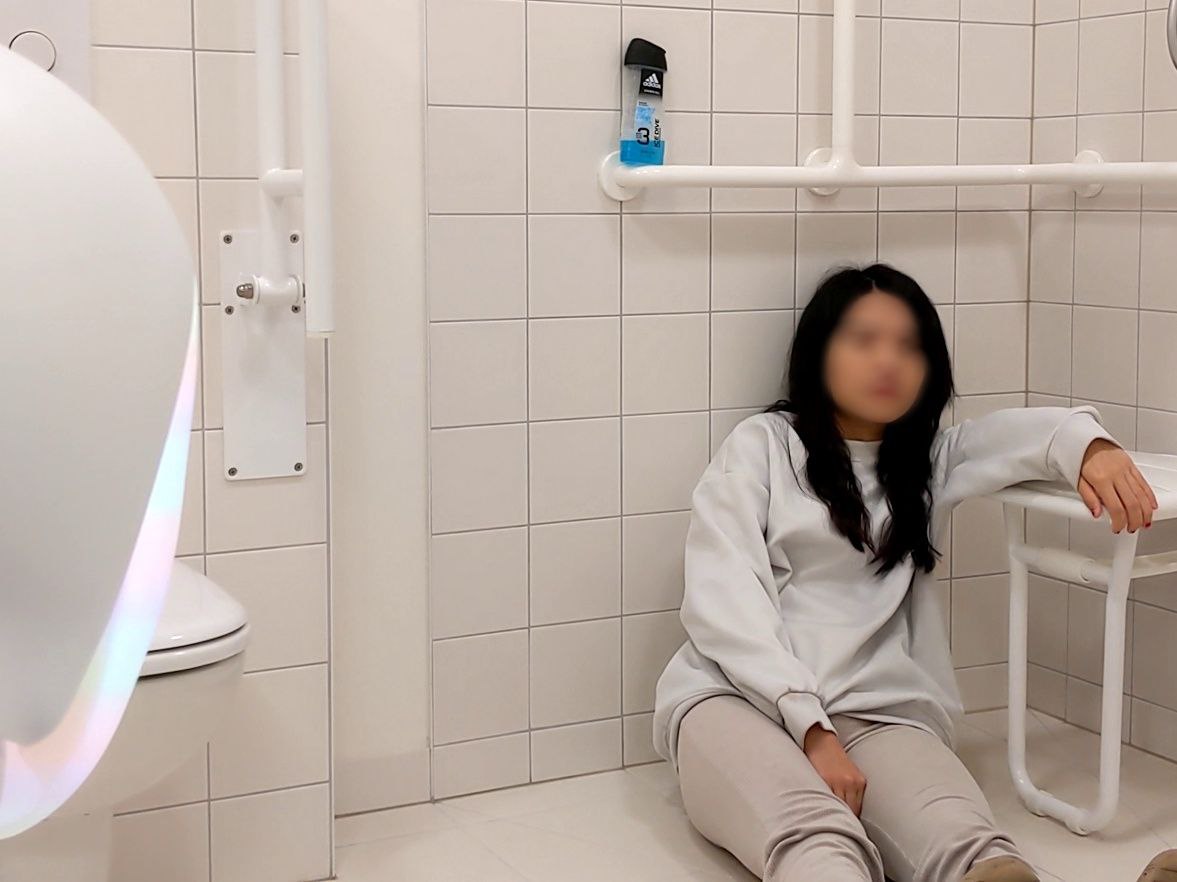}
\includegraphics[width=0.45\linewidth]{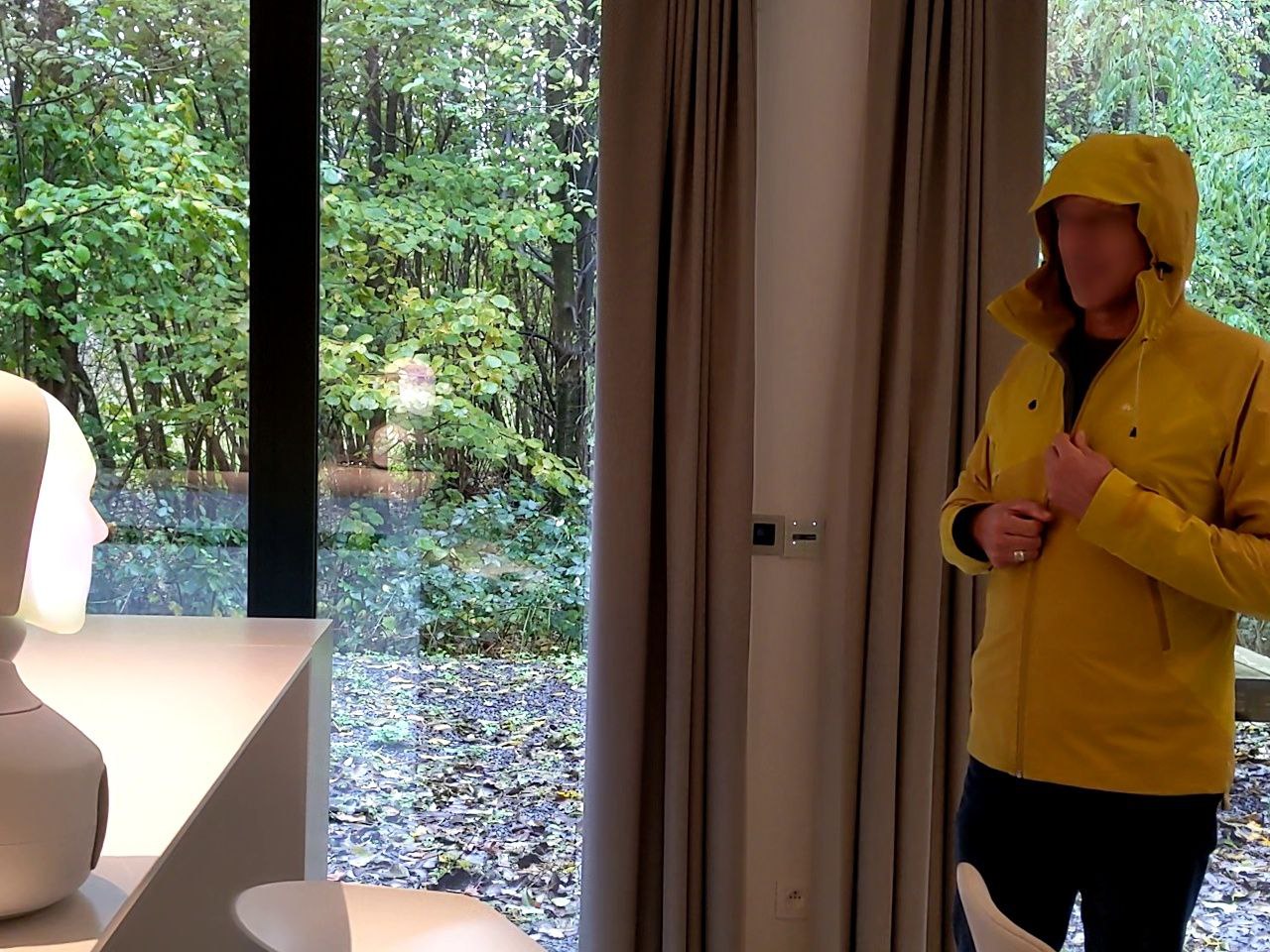}

\caption{\label{fig:interactions}Frames from the videos of the interactions showing the kitchen, bedroom, bathroom and entrance environments.}
\end{figure}

To demonstrate the capabilities and the results obtained, we report and discuss the settings and main highlights of six interactive sessions.
For these sessions, we used the system previously described to empower a Furhat robot with vision and dialogue capabilities.
Since the dialogues generated are highly context-dependent, we ran the sessions in five different environments (see Figure~\ref{fig:interactions}):
a lab, a kitchen, the home entrance, a bathroom, and a bedroom.

Session 1: in a lab with desks and a window; it is dark outside.
The system recognised the environment correctly, identifying that the user was in a lab or a workplace.
It then asks the user if he is working on something interesting, and recommends not to work late hours even if the project is exciting.
Unexpectedly, the system deduced that it was late, probably from the dark windows.

Session 2: in a kitchen, in front of a counter.
The system recognises that the person in front of it is cooking, and when asked is able to come up with suggestions on what to prepare.
This setting is probably one of the most realistic use cases for a vision-enabled assistant in the home and showcases the intrinsic knowledge contained in LLM.
Provided a higher frame rate, we can imagine the system being able to follow the actions of the user and guiding her step-by-step through countless recipes.

Session 3: in a kitchen, in front of a coffee machine.
In this session, the user opens the conversation with a direct question: ``Hi, can you help me with this?''.
The system is able to recognise that the appliance in question is the coffee machine, and provides detailed instructions on how to use it.
This example shows that the LLMs are able to disambiguate the user's request, without the need of providing additional information.

Session 4: in the home entrance; wearing a rain jacket.
In this case, the attention of the robot is immediately attracted by the bright-coloured rain jacket.
The robot asks whether it is raining outside and proceeds to have a conversation about the weather.
More and more frequently conversational agents and social robots are used for entertaining and keeping company, improving the well-being of isolated people.
This session is an example of how much more engaging conversation with these systems can be when powered by images together with text.

Session 5: in a bathroom; a person is lying on the ground.
The person in the shot starts the interaction asking for help.
As expected, the extremely rational response of the system and the calm voice of the speech synthesizer are in contrast with the criticality of the moment.
However, what is relevant is that the system is able to understand that the situation is problematic, and offers advice on how to solve the problem, fully knowing the limitations of its capabilities.

Session 6: in a bedroom; holding a jacket and a t-shirt.
The user tells the robot that it is raining and she has to choose what to wear.
The robot is able to recognise that the person is undecided between the two pieces of clothes held, and sees that the jacket has a hood.
It then proceeds to suggest to wear the jacket and keep the other to stay inside.

Additional observations: no frames and dark images.
In one instance a technical problem produced a prompt without frames.
The system suggested that there might be some problems since it was not receiving any images.
While this was ultimately true, and the response helped in noticing the issue quickly, it was not the desired effect, because the system in a sense broke out of character.
In another case, an object obstructing the camera view caused black frames to be sent to the system.
This time the system did not break out of character and answered: ``Hello, it looks like you are in a pitch-dark room. I cannot see anything. Maybe you can turn on the lights?''.
We think that these cases show the flexibility of the system and how it uses the power of the LLM to adapt to edge cases and unusual situations.

\subsection{Regarding the Measures Identified}

Considering the measures of effective conversational interactions previously identified, we want to discuss how the system performs in each area.
Currently, the system does not have memory of previous interactions, so it is not able to customise the conversation to the user, but this will be addressed in future work.
Active listening capabilities are notable, with the system responding thoughtfully to the user, showing nuanced understanding and genuine interest.

In terms of adaptability, the system excels in providing relevant information across diverse scenarios, such as suggesting recipes or offering guidance on appliance usage.
The LLMs that power the system excel in adapting to unexpected inputs and generating coherent responses, ignoring irrelevant or inaccurate information.
However, the system is not suitable for goal-oriented tasks, as it can get carried away with the conversation and lose track of the original goal.

The new vision capabilities of LLMs greatly improve context awareness, and a careful prompting strategy ensures that the system is able to move the conversation forward with salient responses without getting lost in describing the image details.
The conversations with the robot seem engaging and entertaining, although this has to be verified through a rigorous evaluation.
Injecting elements of entertainment and humour into responses emerges as an area for potential enhancement.

\subsection{Ablation Experiments and Future Works}

We conducted a number of preliminary ablation trials.
We have found that reducing the number of frames to one every five seconds still leads to satisfying results.
However, sending frames only when the user interacts with the system caused a decrease in the quality of the responses, which were extremely focused on the image.
Further and more structured experiments are needed to determine which are the best techniques and frame rates to use.

In addition, we tried to reduce the resolution of the frames, to reduce the prompt size.
To maintain the most details out of the visual input, the last image in the prompt is kept at full resolution and is scaled down when a new one is inserted.
Although again we do not have quantitative measures to support our thesis, we think that the quality of the responses is not decreased by this technique.

Our implementation has two big limitations.
The first one is the response speed.
We have noticed that in the morning (time zone: UTC+1) the system is perfectly usable, with responses' speed in the order of 1 second.
However, already approaching noon, the system slows down noticeably, with responses that can lag up to 7 seconds.
We have observed this behaviour over multiple days and we suppose it has to do with the increased loads on the servers that host the LLM we use.
Another gain in speed could be obtained by adopting more performant transformer-based speech-to-text techniques, which could also support multiple languages.

The second limitation is the temporal resolution.
Currently, the frame rate cannot be fast enough to allow the model to pick up gestures.
While this could be solved in the future by improvements in the LLMs' elaboration speed, another possible solution is empowering the model to access, when it deems it necessary, a memory of frames that are not in the prompt, allowing it to \emph{look back} and gather more details from the input.

\section{Conclusion}

The implementation of a vision-enabled dialogue system marks a significant advancement in the realm of conversational agents.
The integration of visual information, captured in real-time frames from a video feed, empowers the system with an increased awareness of its surroundings.
As a consequence, this fusion of language and vision facilitates a more contextually aware and immersive conversational experience.

One key strength of the proposed system lies in its ability to handle both textual and visual input seamlessly, without a predefined ordering.
The conversation manager effectively manages the dialogue history and visual frames, creating a prompt that captures the essential elements of the ongoing interaction.
The system's prompt engineering plays a pivotal role in maintaining context while mitigating the challenges posed by an increasing amount of data.
The summarisation of frames ensures a balance between context awareness and computational efficiency.

However, it is essential to acknowledge the challenges inherent in implementing vision-enabled dialogue systems.
The balance between textual and visual information is crucial, as the system must avoid getting sidetracked by images and losing the thread of the conversation.
Moreover, the computational demands associated with processing visual data, especially in real-time, pose challenges that need careful consideration.
This leads to two issues in the current implementation: the low frame rate does not allow capturing gestures, and the response time can be too slow in some instances.
While we have struck a good balance in this regard, and have obtained stunning results using a Furhat robot, future improvements could explore techniques to optimise the system's efficiency further.

In conclusion, the vision-enabled dialogue system presented represents a step forward in creating more immersive and context-aware conversational agents.
The successful integration of visual cues enhances the system's understanding of the world, opening new avenues for applications where a combination of textual and visual information is essential.

\section*{Acknowledgements}

Funded by the Horizon Europe VALAWAI project (grant agreement number 101070930).

\printbibliography
\end{document}